\documentclass{article}
\usepackage{amsmath}
\usepackage{amssymb}
\usepackage{tikz}

\usepackage{kr}

\usepackage{times}
\usepackage{soul}
\usepackage{url}
\usepackage[hidelinks]{hyperref}
\usepackage[utf8]{inputenc}
\usepackage[small]{caption}
\usepackage{graphicx}
\usepackage{amsmath}
\usepackage{amsthm}
\usepackage{booktabs}
\usepackage{algorithm}
\usepackage{algorithmic}
\title{Layer-of-Thoughts Prompting (LoT): Leveraging LLM-Based Retrieval with Constraint Hierarchies}

\author{%
Wachara Fungwacharakorn\thanks{These authors contributed equally}\and
Nguyen Ha Thanh\footnotemark[1]\and
May Myo Zin\footnotemark[1]\and
Ken Satoh\footnotemark[1] \\
\affiliations
Center of Juris-Informatics, ROIS-DS, Tokyo, Japan
\emails
\{wacharaf, maymyozin, nguyenhathanh, ksatoh\}@nii.ac.jp
}

\begin{document}

\maketitle

\begin{abstract}
This paper presents a novel approach termed Layer-of-Thoughts Prompting (LoT), which utilizes constraint hierarchies to filter and refine candidate responses to a given query. By integrating these constraints, our method enables a structured retrieval process that enhances explainability and automation. Existing methods have explored various prompting techniques but often present overly generalized frameworks without delving into the nuances of prompts in multi-turn interactions. Our work addresses this gap by focusing on the hierarchical relationships among prompts. We demonstrate that the efficacy of thought hierarchy plays a critical role in developing efficient and interpretable retrieval algorithms. Leveraging Large Language Models (LLMs), LoT significantly improves the accuracy and comprehensibility of information retrieval tasks.
\end{abstract}

\section{Introduction}
% \subsection{Background}
% \subsection{Motivation}
In recent years, there has been an explosion of interest in various prompting techniques for leveraging Large Language Models (LLMs). However, many existing works provide frameworks that are overly general, lacking the specificity needed to tackle detailed prompt-related challenges. One critical aspect that has been largely overlooked is the differentiation between prompts in multi-turn interactions. 

Our paper seeks to fill this gap by examining the importance of hierarchical relationships among prompts. Specifically, we introduce the concept of thought hierarchies to filter and refine candidate responses, creating more structured and explainable retrieval processes. We believe that the strength of thought hierarchy is a pivotal factor in designing algorithms that are both efficient and interpretable. By incorporating constraint hierarchies within LoT, our approach not only enhances retrieval accuracy but also addresses the scalability and comprehension issues associated with complex queries.

% \subsection{Overview of Layer of Thoughts Prompting (LoT)}
The LoT framework extends the Graph-of-Thoughts by representing reasoning as a graph where nodes, called \emph{thoughts}, denote reasoning steps. These thoughts are partitioned into layers, and categorized into \textbf{layer thoughts}, which handle conceptual step given by the user, and \textbf{option thoughts}, which assist in finding solutions.

The process starts with initializing layer thoughts, and proceeds from the first layer down to the last. Each layer thought receives input from the prior layer and branches into option thoughts, generating partial solutions. Aggregated outputs from option thoughts are passed to the next layer or used to expand thoughts dynamically until the last layer is reached.

For retrieval tasks, the LoT framework utilizes hierarchical levels to filter and rank documents from a given corpus based on a query. Relevance scores can be aggregated using several metrics, ensuring efficient and effective document ranking. Outputs at each layer progressively filter documents, providing clear explanations based on aggregated scores to justify document relevance.

\section{Related Work}
\subsection{Traditional Information Retrieval}
Before neural networks became widely popular, classical NLP approaches were the go-to methods for solving information retrieval tasks \cite{cooper1971definition,luhn1957statistical,salton1988term}. These techniques primarily employed various lexical matching strategies. The researchers proposed both logical and statistical models to measure the similarity between queries and potential matches. Despite their advantages—such as fast computation and versatility—these approaches mainly relied on text morphology. Since morphological similarity does not necessarily equate to semantic similarity, achieving high accuracy in semantic similarity posed a challenge. As a result, the performance of these methods is limited, particularly in cases where document-query pairs exhibit non-overlapping text but semantic relevance, or overlapping text lacking semantic relation.

\subsection{Neural Information Retrieval}
% \subsection{Constraint-Based Retrieval} Thanh: I think every approach is constraint-based 
Early pre-trained neural network models, including pre-trained word embeddings such as Word2Vec~\cite{mikolov2013distributed}, GloVe~\cite{pennington2014glove}, and FastText~\cite{mikolov2018advances}, have proven highly effective in capturing semantic relationships between words. In the legal domain, a significant advancement is Law2Vec \cite{chalkidis2019deep}, a specialized word embedding trained on legal texts, which has demonstrated substantial effectiveness.

More recently, Transformer-based models \cite{vaswani2017attention} have established new performance benchmarks across various tasks. Notable examples include BERT~\cite{devlin2018bert}, BART~\cite{lewis-etal-2020-bart}, the GPT series~\cite{radford2018improving,radford2019language}, and Sentence-BERT~\cite{reimers-2019-sentence-bert}, which have achieved state-of-the-art results in general applications. In legal text analysis, these advancements have been mirrored by models such as Legal-BERT \cite{yilmaz2019applying,10.1007/978-3-030-79942-7_13,yoshioka2021bert,nguyen2022transformer}.
Nguyen et al. suggest using attentive neural network-based text representation for statute law document retrieval, showing effective results in this area \cite{nguyen2022attentive}. 

\subsection{Reasoning Topologies for LLM}
% (X-of-Thoughts)
The field of LLM prompting has seen a surge in interest regarding reasoning topologies, also known as X-of-Thoughts  \cite{besta2024demystifying,liu-etal-2023-plan}. A seminal work in this area is Chain-of-Thoughts \cite{wei2022chain}, which instructs LLMs to not only generate an output but also reveal their step-by-step reasoning process (each step is called a \emph{thought}). This concept was extended to Tree-of-Thoughts \cite{yao2024tree}, where LLMs can branch into multiple potential thoughts after each step and evaluate their progress towards the solution. Subsequently, Graph-of-Thoughts \cite{besta2024got} generalized this further, allowing each thought to connect with others. Connections can involve aggregating outputs from other thoughts or refining the thought itself. Graph-of-Thoughts presents a model to formalize topological prompting using four components:

%Besides graphs, Program-of-Thoughts \cite{chen2022program} explores alternative structures for representing reasoning steps using programs. Some X-of-Thoughts focus on augmenting thoughts and reducing computational overhead, such as Skeleton-of-Thoughts \cite{ning2024skeleton}, Algorithm-of-Thoughts \cite{sel2023algorithm}, and Buffer-of-Thoughts \cite{yang2024buffer}.

% \wnote{Please feel free to rewrite this section as I saw Thanh-san have some reviews regarding this section in the slides.} Thanh: I think your literature review is adequate 

%\subsection{Graph-of-Thoughts}
%The framework in this paper is extended from Graph-of-Thoughts \cite{besta2024got} so we recap it as follows. Graph-of-Thoughts presents a model to formalize topological prompting using four components:

\begin{enumerate}
    \item \textbf{Reasoning process}: The reasoning process is represented as a directed graph. Each node in the graph represents one reasoning step producing a piece of information towards a solution. A direct edge from one thought to another represent that an output (the piece of information) from the start node is considered as an input of the end node.
    \item \textbf{Thought transformation}: The thought transformation is a transition function (probably involving LLMs) that gives next plausible reasoning processes from the current one.  The function includes adding new thoughts and edges based on the existing thought (e.g., branching or aggregating) or adding new edges without adding new thoughts (e.g., refining).
    \item \textbf{Evaluation function}: This function evaluates thoughts in the reasoning progress as a score towards the solution, and LLMs may involve in this function. Sometimes, an evaluation function needs to consider the whole reasoning process because a score of one thought may be relative to others.
    \item \textbf{Ranking function}: This function ranks thoughts to consider a sole output for the main task. This function is related to the evaluation function as it mostly rank thoughts according to the scores.
\end{enumerate}

\subsection{Constraint Hierarchies}
%Let $D$ be the domain of variables.  Let $X$ be the set of all variables. A valuation is a mapping $\theta : X \mapsto D$. Let $\Theta$ be the set of all valuations. Let $C$ be a set of constraints. Constraint hierarchies consider two types of constraints: \emph{hard constraints} (or \emph{required constraints}) and \emph{soft constraints} (or \emph{preferential constraints}).

LoT Prompting in this paper is inspired by constraint hierarchies \cite{borning1992constraint}, which are used for modelling constraints with strengths. In constraint hierarchies, each constraint is evaluated by an error function $e(c,\theta)$ that returns a non-negative real number indicating how nearly constraint $c$ is satisfied for a valuation $\theta$; and the function returns a zero value when the $c$ is exactly satisfied. The strengths of the constraints are indicated by a non-negative integer. Conventionally, constraints with strength level $0$ are hard constraints (or \emph{required constraints}) and constraints with strength level $i \geq 1$ are soft constraints (or \emph{preferential constraints}) where a higher value of the strength indicates that the constraint is weaker. Following this, we can represent a constraint hierarchy $H$ as a partition $H_0, H_1, \ldots, H_\ell$ where each $H_i$, called a \emph{level}, contains all of the constraints in $H$ with strength $i$. 

Solutions to constraint hierarchies must satisfy all hard constraints and satisfy soft constraints as much as possible. That is, there is no valuation outside the solutions that better satisfies the soft constraints and satisfies all of the hard constraints. To consider which valuation better satisfies the soft constraints, constraint hierarchies introduce comparators, which are later extended into level comparators and hierarchical comparators \cite{hosobe2003foundation}. 
Given valuations $\theta,\theta', \theta''$, a level $H_i$ of a constraint hierarchy $H$ with strength $i$, the level comparator $\preceq_i$ must satisfy the following conditions

\begin{enumerate}
    \item If $e(c,\theta) = e(c,\theta'')$ for every $c \in H_i$, then $\theta \preceq_i \theta'$ if and only if $\theta'' \preceq_i \theta'$
    \item If $e(c,\theta') = e(c,\theta'')$ for every $c \in H_i$, then $\theta \preceq_i \theta'$ if and only if $\theta \preceq_i \theta''$
    \item If $e(c,\theta) \leq e(c,\theta')$ for every $c \in H_i$, then $\theta \preceq_{i} \theta'$
    \item If $\theta \preceq_{i} \theta'$ and $\theta' \preceq_{i} \theta''$, then $\theta \preceq_{i} \theta''$
\end{enumerate}

    The first and second conditions state that if errors after applying valuations are the same for every constraint in the level, then the level comparator works the same. The third condition states that if errors after applying one valuation is less than or equal to the errors after applying another for every constraint in the level, then the former is better than or equal to the latter.  The forth condition indicates the transitive property of the level comparator (we omit one condition from \cite{hosobe2003foundation} as we consider only one constraint hierarchy each time).
For a level comparator $\preceq_{i}$, we write
\begin{itemize}
    \item $\theta \sim_{i} \theta'$ if $\theta \preceq_{i} \theta'$ and $\theta' \preceq_{i} \theta$;
    \item $\theta \prec_{i} \theta'$ if $\theta \preceq_{i} \theta'$ and $\theta' \not\preceq_{i} \theta$; and
    \item $\theta \not\sim_{i} \theta'$ if $\theta \not\preceq_{i} \theta'$ and $\theta' \not\preceq_{i} \theta$.
\end{itemize}

If $\theta \not\sim_{i} \theta'$ is not possible (meaning that, we have either $\theta \sim_{i} \theta'$, $\theta \prec_{i} \theta'$, or $\theta' \prec_{i} \theta$), $\preceq_{i}$ is said to be \emph{total}. If $\theta \preceq_{i} \theta'$ implies that $e(c,\theta) \leq e(c,\theta')$ for every $c \in H_i$, $\preceq_{i}$ is said to be \emph{local}. A local $\preceq_{i}$ is generally not total because there might be a level $H_i$ with two constraints $c_1, c_2$ such that $e(c_1,\theta) > e(c_1,\theta')$ and $e(c_2,\theta) < e(c_2,\theta')$ and hence $\theta \not\sim_{i} \theta'$. If there is a numerical aggregation function $g(\theta,H_i)$ such that  $g(\theta,H_i) \leq g(\theta',H_i)$ if and only if $\theta \preceq_{i} \theta'$, then $\preceq_{i}$ is said to be \emph{global}. A global $\preceq_{i}$ is always total because a numerical comparator is total.

A hierarchical comparator $\prec$ covers overall comparisons by considering level by level, that is $\theta \prec \theta'$ if there exists $k \in \{1,\ldots,\ell\}$ such that $\theta \prec_{k} \theta'$ and for $i < k$, $\theta \sim_{i} \theta'$. Specific types of hierarchical comparators can be defined based on types of level comparators and aggregation functions used. Given the weight for constraint $c$ is denoted by $w_c$, the original paper of constraint hierarchies \cite{borning1992constraint} provides some examples of (hierarchical) comparators as follows.
\begin{itemize}
    \item \emph{locally-better} is a hierarchical comparator where a local-level comparator is used.
    \item \emph{weighted-sum-better} is a hierarchical comparator with an aggregation function  $g(\theta,H_i) = \Sigma_{c \in H_i} w_c e(c,\theta)$.
    \item \emph{worst-case-better} is a hierarchical comparator with an aggregation function  $g(\theta,H_i) = max(\{w_c e(c,\theta)|c \in H_i\})$.
    \item \emph{least-squares-better} is a hierarchical comparator with an aggregation function  $g(\theta,H_i) = \Sigma_{c \in H_i} w_c e(c,\theta)^2$.
\end{itemize}

\section{LoT Prompting}
% \subsection{Framework Design}
% \subsection{Constraint Hierarchies}
% \subsection{Integration with LLMs}
% \subsection{Algorithm for Retrieval Process}

\begin{figure}
    \centering
    \scalebox{0.7}{
        \input{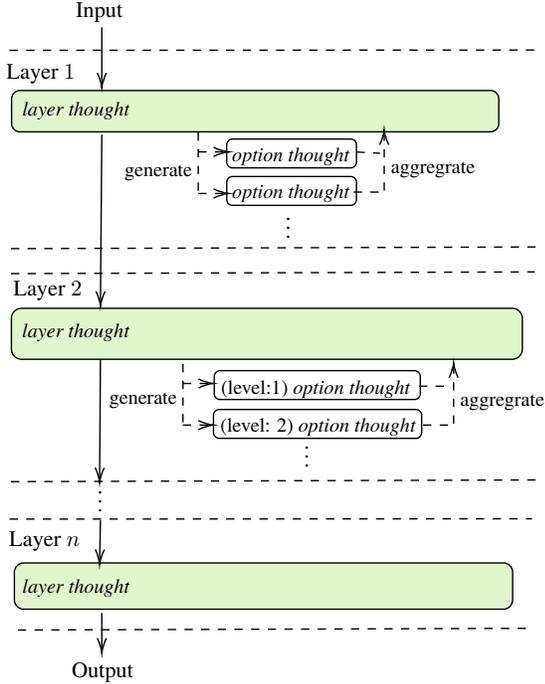}
    }
    \caption{LoT Prompting}
    \label{fig:diagram}
\end{figure}

Figure \ref{fig:diagram} illustrates Layer-of-Thoughts (LoT) prompting. The approach extends Graph-of-Thoughts \cite{besta2024got}, which represents the reasoning process as a graph. The graph contains nodes, called \emph{thoughts}, representing reasoning steps.  Each thought receives outputs from the previous thoughts, and utilizes them to instruct LLMs in generating a partial solution for the next thoughts. In LoT Prompting, each thought is assigned a number to show which \emph{layer} they are in. There are two types of thoughts in the graph.

\begin{itemize}
    
    \item \textbf{Layer thought}:
    Each layer contains a layer thought to handle conceptual step given by the user. A layer thought in layer $i$ has a previous layer thought in layer $i-1$ and a next layer thought in layer $i+1$, except the first and the last layer thoughts. A layer thought has a role to receive outputs from the previous layer thought, determine generating option thoughts, aggregate outputs from the generated option thoughts, and forward the aggregated output to the next layer thought.

    \item \textbf{Option thought}:
    Each layer contains multiple option thoughts or even none of them (e.g., the last layer in Figure \ref{fig:diagram}). An option thought in layer $i$ receives the inputs from the layer thought in the same layer, and generates a partial solution as the output for aggregating in the layer thought. A layer thought may generate equal option thoughts (e.g., Layer $1$ in Figure \ref{fig:diagram}) or prioritize option thoughts into several \emph{levels}, defined by a positive integer (e.g., Layer $2$ in Figure \ref{fig:diagram}). We consider a layer with equal option thoughts as \emph{single-level} layer and a layer with prioritized option thoughts as \emph{multiple-level} layer
\end{itemize}

Thoughts in LoT Prompting are initiated as follows. 

\begin{enumerate}
    \item A user designs conceptual steps for the task. 
    \item Starting from the first step, each step initiates a new layer thought. The layer thought receives inputs from the previous layer thought (or from the main task for the first layer thought), and instructs LLMs to suggest appropriate criteria for that step. The criteria can be generated equally or prioritized in order of importance.
    \item For each criterion, an option thought is generated. Each option thought instructs LLMs or performs calculation, and then provides a partial solution for the task according to the criterion.
    \item The layer thought then aggregates the outputs from option thoughts in the layer, and determines whether the aggregated output is appropriate. If it is appropriate, then the output is fed to the new layer thought (invoke Step 2 for initiating a new layer thought from the next step). If it is not appropriate (e.g., no inputs from the previous layer pass the criteria), we can instruct LLMs to refine the criteria and initiate new option thoughts again (backtrack to Step 3).
\end{enumerate}

\subsection{Retrieval Tasks}

    In this paper, we explore leveraging LoT Prompting for retrieval tasks. Given a query $q$ and a document corpus $D$, retrieval tasks aim to find the most relevant documents in $D$ with respect to the query $q$. Retrieval tasks are usually evaluated using relevance scores. In LoT Prompting, we assume that each option thought can be evaluated using a non-negative relevance score $\rho(d,t)$ of a document $d \in D$ for an option thought $t$ (greater score indicates more relevance). A relevance function is \emph{binary} if the score can only be either $0$ or $1$. That is, the function returns $1$ if the document $d$ passes the criterion corresponding to the thought $t$ and returns $0$ if it fails. 
        
    We suggest several metrics to aggregate outputs from option thoughts in a single-level layer as follows. 
    \begin{itemize}
        \item \texttt{all}:  In this metric, relevance scores of documents in the aggregated output must be greater than $0$ for every option thought within this layer. This is inspired from hard constraints in constraint hierarchies.
        
        \item \texttt{at-least-k} (where $k$ is a positive integer): In this metric, relevance scores of documents in the aggregated output must be greater than $0$ for at least $k$ option thoughts within this layer. This metric is relaxed from \texttt{all}.
       
        \item \texttt{locally-better}: 
        This metric is a relative metric, inspired by \emph{locally-better} in constraint hierarchies. Some studies refer to this metric as Pareto efficiency. In this metric, the aggregated output contains only the top-ranked documents from the input, where no input document surpasses the relevance score of the top-ranked document for every option thought within this layer.
        
        \item \texttt{max-count}: Since \texttt{locally-better} is not a total order, \texttt{max-count-better} is a total order relaxed from that metric. \texttt{max-count-better} can only be used for binary relevance functions. In this metric, the aggregated output contains documents that maximize the number of passed criteria.
        
        \item \texttt{max-weight}: This metric is inspired by \emph{weight-sum-better} in constraint hierarchies. Given 
        the weight for the thought $t$ as $w_t$, the aggregated output contain documents that maximize $\Sigma_{t \in T^i} w_t \rho(d,t)$ where $T^i$ be the set of all of option thoughts in the layer $i$.
    \end{itemize}

    For \texttt{max-count} and \texttt{max-weight}, they do not need to restrict the aggregated output for only documents that maximize such metrics, but the layer thought can present scores of each input document for ranking instead. We call them \texttt{rank-count} and \texttt{rank-weight} respectively. These metrics can be extended for a multiple-level layer by aggregating outputs from the strongest level first and weaker levels successively. 

    One goal of LoT prompting is to reduce computations by skipping the need to explore the entire corpus with all option thoughts. In LoT prompting, option thoughts are partitioned by layers with layer thoughts that determine if further exploration is necessary. Let $T$ denote the set of all option thoughts, $n$ denote the total number of layers, $m_i$ denote the number of levels in layer $i$ ($m_i$ can vary across layers), and $T_i$ denote the set of all option thoughts in layer $i$. A \emph{hierarchy} of $T^i$ refers to a tuple $\langle T^i_1, \ldots, T^i_{m_i} \rangle$, where $T^i_j$ contains all of the option thoughts in the layer $i$ with level $j$. $T$ can be represented as a nested hierarchy: 
    
    \begin{center}
        $\langle \langle T^1_1, \ldots, T^1_{m_1} \rangle, \langle T^2_1, \ldots, T^2_{m_2} \rangle, \ldots, \langle T^n_1, \ldots, T^n_{m_n} \rangle \rangle$
    \end{center}

    It can be observed that the multiple-layer prompting has the same expessive power as a single layer with multiple levels because the nested hierarchy works in the same way as the flatten hierarchy:

    \begin{center}
        $\langle T^1_1, \ldots, T^1_{m_1}, T^2_1, \ldots, T^2_{m_2}, \ldots,  T^n_1, \ldots, T^n_{m_n} \rangle$
    \end{center}
    
    This indicates a structured arrangement of thoughts, with layers representing distinct blocks of thoughts to convey semantic understanding, and levels representing more specific evaluation criteria. The observation helps simplifying the explanation of the reasoning process. For example, assuming we explain an output with a set $E = \{t \in T~|~ \rho(d,t) > 0\}$. Consequently, $E$ can be represented as $\langle E^1_1, \ldots, E^1_{m_1}, E^2_1, \ldots, E^2_{m_2} , \ldots,  E^n_1, \ldots, E^n_{m_n} \rangle$, where $E^i_j$ contains all of the option thoughts in $E$ in the layer $i$ with level $j$.

\section{Application-Driven Experimental Setup}
In this section, we present two experimental applications corresponding to two specific use cases: Japanese Civil Law retrieval and normative sentence retrieval. Through these applications, we verify the efficacy of the Layer-of-thought approach and identify its limitations.
\subsection{Japanese Civil Law Retrieval}
Legal AI is a niche field that leverages computer and artificial intelligence techniques to make legal processes more efficient and effective. The rapid progression of AI-based tools is set to emancipate legal professionals from cumbersome tasks such as document searches and contract reviews. The COLIEE competition \cite{goebel2024overview} is held annually to promote research in legal information processing. This competition includes various challenges like document retrieval and legal entailment.

Participants are provided with a legal question $Q$ from the Japanese Bar Exam and are tasked with retrieving relevant articles $A1, A_2, \dots, A_n$ from the Japanese Civil Code. The ability to retrieve pertinent legal articles is crucial for several reasons. It enables legal professionals to quickly access relevant legal precedents and statutes, thereby enhancing their decision-making process. However, the task is particularly challenging due to the complex language and vast amount of information contained within legal texts.

The Japanese Civil Code is rich with intricate, specialized terminology, making the retrieval task demanding. Moreover, the volume of information makes it difficult to pinpoint the exact articles that are relevant to a given legal question. These challenges highlight the importance and difficulty of the task, making it a suitable candidate to test the feasibility of the Layer-of-Thoughts approach.

The evaluation metrics for this task include the macro average of F2, precision, and recall. This comprehensive set of metrics ensures a robust assessment of each team’s performance. The calculations for these measures are as follows:
\begin{equation*}
\text{Precision} = \text{avg} \left( \frac{\text{\# correct articles}}{\text{\# retrieved articles}} \right)
\end{equation*}
\begin{equation*}
\text{Recall} = \text{avg} \left( \frac{\text{\# correct articles}}{\text{\# relevant articles}} \right)
\end{equation*}
\begin{equation*}
\text{F2} = \text{avg} \left( \frac{5 \times \text{Precision} \times \text{Recall}}{4 \times \text{Precision} + \text{Recall}} \right)
\end{equation*}

\begin{figure}
    \centering
    \scalebox{0.7}{
        \tikzset{every picture/.style={line width=0.75pt}, every node/.style={scale=1.2,yshift=-2}} %set default line width to 0.75pt      

\begin{tikzpicture}[x=0.75pt,y=0.75pt,yscale=-1,xscale=1]
%uncomment if require: \path (0,526); %set diagram left start at 0, and has height of 526

%Straight Lines [id:da28324363039045974] 
\draw    (29.47,21.54) -- (29.47,66.16) ;
\draw [shift={(29.47,68.16)}, rotate = 270] [color={rgb, 255:red, 0; green, 0; blue, 0 }  ][line width=0.75]    (10.93,-3.29) .. controls (6.95,-1.4) and (3.31,-0.3) .. (0,0) .. controls (3.31,0.3) and (6.95,1.4) .. (10.93,3.29)   ;
%Rounded Rect [id:dp6072579570561454] 
\draw   (44.47,111.12) .. controls (44.47,108.82) and (46.33,106.96) .. (48.63,106.96) -- (334.36,106.96) .. controls (336.65,106.96) and (338.52,108.82) .. (338.52,111.12) -- (338.52,123.6) .. controls (338.52,125.9) and (336.65,127.76) .. (334.36,127.76) -- (48.63,127.76) .. controls (46.33,127.76) and (44.47,125.9) .. (44.47,123.6) -- cycle ;
%Straight Lines [id:da9879616616522671] 
\draw  [dash pattern={on 4.5pt off 4.5pt}]  (12,48) -- (402.52,48) ;
%Straight Lines [id:da2943452340851982] 
\draw  [dash pattern={on 4.5pt off 4.5pt}]  (12,184) -- (410.52,183.36) ;
%Straight Lines [id:da3385350560335756] 
\draw    (31.47,101.76) -- (30.48,223.76) ;
\draw [shift={(30.47,225.76)}, rotate = 270.46] [color={rgb, 255:red, 0; green, 0; blue, 0 }  ][line width=0.75]    (10.93,-3.29) .. controls (6.95,-1.4) and (3.31,-0.3) .. (0,0) .. controls (3.31,0.3) and (6.95,1.4) .. (10.93,3.29)   ;
%Straight Lines [id:da1903812936389928] 
\draw  [dash pattern={on 4.5pt off 4.5pt}]  (12,202) -- (410.52,202.36) ;
%Straight Lines [id:da10101858040408973] 
\draw  [dash pattern={on 4.5pt off 4.5pt}]  (12,352) -- (412.52,352) ;
%Straight Lines [id:da37896517120185513] 
\draw    (31.47,259.76) -- (30.48,403.76) ;
\draw [shift={(30.47,405.76)}, rotate = 270.39] [color={rgb, 255:red, 0; green, 0; blue, 0 }  ][line width=0.75]    (10.93,-3.29) .. controls (6.95,-1.4) and (3.31,-0.3) .. (0,0) .. controls (3.31,0.3) and (6.95,1.4) .. (10.93,3.29)   ;
%Rounded Rect [id:dp32255630995685713] 
\draw  [fill={rgb, 255:red, 184; green, 233; blue, 134 }  ,fill opacity=0.4 ] (14.47,417.27) .. controls (14.47,413.59) and (17.45,410.6) .. (21.14,410.6) -- (406.85,410.6) .. controls (410.53,410.6) and (413.52,413.59) .. (413.52,417.27) -- (413.52,437.29) .. controls (413.52,440.97) and (410.53,443.96) .. (406.85,443.96) -- (21.14,443.96) .. controls (17.45,443.96) and (14.47,440.97) .. (14.47,437.29) -- cycle ;
%Straight Lines [id:da48265889665209105] 
\draw  [dash pattern={on 4.5pt off 4.5pt}]  (11,379) -- (409.52,379) ;
%Straight Lines [id:da494004409490133] 
\draw  [dash pattern={on 4.5pt off 4.5pt}]  (17,459) -- (409.52,459) ;
%Straight Lines [id:da32622234080161006] 
\draw    (30.47,446.69) -- (30.51,468.18) ;
\draw [shift={(30.52,470.18)}, rotate = 269.88] [color={rgb, 255:red, 0; green, 0; blue, 0 }  ][line width=0.75]    (10.93,-3.29) .. controls (6.95,-1.4) and (3.31,-0.3) .. (0,0) .. controls (3.31,0.3) and (6.95,1.4) .. (10.93,3.29)   ;
%Rounded Rect [id:dp0932946556619425] 
\draw   (43.47,137.12) .. controls (43.47,134.82) and (45.33,132.96) .. (47.63,132.96) -- (334.36,132.96) .. controls (336.65,132.96) and (338.52,134.82) .. (338.52,137.12) -- (338.52,149.6) .. controls (338.52,151.9) and (336.65,153.76) .. (334.36,153.76) -- (47.63,153.76) .. controls (45.33,153.76) and (43.47,151.9) .. (43.47,149.6) -- cycle ;
%Rounded Rect [id:dp18182133768573028] 
\draw  [fill={rgb, 255:red, 184; green, 233; blue, 134 }  ,fill opacity=0.4 ] (14.33,234.26) .. controls (14.33,230.79) and (17.14,227.98) .. (20.6,227.98) -- (408.24,227.98) .. controls (411.71,227.98) and (414.52,230.79) .. (414.52,234.26) -- (414.52,253.08) .. controls (414.52,256.55) and (411.71,259.36) .. (408.24,259.36) -- (20.6,259.36) .. controls (17.14,259.36) and (14.33,256.55) .. (14.33,253.08) -- cycle ;
%Rounded Rect [id:dp608546828121421] 
\draw   (46.47,277.12) .. controls (46.47,274.82) and (48.33,272.96) .. (50.63,272.96) -- (370.36,272.96) .. controls (372.65,272.96) and (374.52,274.82) .. (374.52,277.12) -- (374.52,289.6) .. controls (374.52,291.9) and (372.65,293.76) .. (370.36,293.76) -- (50.63,293.76) .. controls (48.33,293.76) and (46.47,291.9) .. (46.47,289.6) -- cycle ;
%Rounded Rect [id:dp966206654936995] 
\draw   (45.47,304.12) .. controls (45.47,301.82) and (47.33,299.96) .. (49.63,299.96) -- (370.36,299.96) .. controls (372.65,299.96) and (374.52,301.82) .. (374.52,304.12) -- (374.52,316.6) .. controls (374.52,318.9) and (372.65,320.76) .. (370.36,320.76) -- (49.63,320.76) .. controls (47.33,320.76) and (45.47,318.9) .. (45.47,316.6) -- cycle ;
%Rounded Rect [id:dp5329334302331934] 
\draw  [fill={rgb, 255:red, 184; green, 233; blue, 134 }  ,fill opacity=0.4 ] (13.47,75.27) .. controls (13.47,71.59) and (16.45,68.6) .. (20.14,68.6) -- (403.85,68.6) .. controls (407.53,68.6) and (410.52,71.59) .. (410.52,75.27) -- (410.52,95.29) .. controls (410.52,98.97) and (407.53,101.96) .. (403.85,101.96) -- (20.14,101.96) .. controls (16.45,101.96) and (13.47,98.97) .. (13.47,95.29) -- cycle ;

% Text Node
\draw (9,2) node [anchor=north west][inner sep=0.75pt]   [align=left] {Input: {\footnotesize \textit{(Example query: John, a 15 year old kid, made a contract ...) }}\\\textit{{\footnotesize 		}}};
% Text Node
\draw (12,463) node [anchor=north west][inner sep=0.75pt]   [align=left] {Output: };
% Text Node
\draw (139,150) node [anchor=north west][inner sep=0.75pt]   [align=left] {$\displaystyle \vdots $};
% Text Node
\draw (246,202) node [anchor=north west][inner sep=0.75pt]  [font=\normalsize] [align=left] {Semantic Filtering Layer};
% Text Node
\draw (158.65,157.63) node [anchor=north west][inner sep=0.75pt]  [font=\footnotesize] [align=left] {(single - level)};
% Text Node
\draw (244,382) node [anchor=north west][inner sep=0.75pt]  [font=\normalsize] [align=left] {Final Confirmation Layer};
% Text Node
\draw (15.14,73.6) node [anchor=north west][inner sep=0.75pt]   [align=left] {{\footnotesize \textit{From the query, please list keywords the article should have }}};
% Text Node
\draw (242,47) node [anchor=north west][inner sep=0.75pt]  [font=\normalsize] [align=left] {Keyword Filtering Layer};
% Text Node
\draw (46.63,107.96) node [anchor=north west][inner sep=0.75pt]  [font=\footnotesize] [align=left] {Does the article contain a word $\displaystyle < minor >$ ?};
% Text Node
\draw (47.63,132.96) node [anchor=north west][inner sep=0.75pt]  [font=\footnotesize] [align=left] {Does an article contain a word $\displaystyle < contract >$ ?};
% Text Node
\draw (14.33,231.26) node [anchor=north west][inner sep=0.75pt]   [align=left] {{\footnotesize \textit{From the query, please list ordered criteria for the relevant articles}}};
% Text Node
\draw (163.65,326.63) node [anchor=north west][inner sep=0.75pt]  [font=\footnotesize] [align=left] {(multiple - level)};
% Text Node
\draw (50.63,272.96) node [anchor=north west][inner sep=0.75pt]  [font=\footnotesize] [align=left] {(level:1) Is the article relevant to $\displaystyle < contract\ law >$ ?};
% Text Node
\draw (49.03,300.96) node [anchor=north west][inner sep=0.75pt]  [font=\footnotesize] [align=left] {(level:2) Is the article relevant to $\displaystyle < juridical\ act >$ ?};
% Text Node
\draw (147,317) node [anchor=north west][inner sep=0.75pt]   [align=left] {$\displaystyle \vdots $};
% Text Node
\draw (14.47,417.27) node [anchor=north west][inner sep=0.75pt]   [align=left] {{\footnotesize \textit{Does the article indeed relevant to the query ?}}};
% Text Node
\draw (13,496) node [anchor=north west][inner sep=0.75pt]   [align=left] {{\footnotesize \textit{* The prompts and examples were simplified for illustration purpose}}};
% Text Node
\draw (53,21) node [anchor=north west][inner sep=0.75pt]   [align=left] {\textit{{\footnotesize + (A set of candidate legal articles)}}};
% Text Node
\draw (69,462.91) node [anchor=north west][inner sep=0.75pt]   [align=left] {{\footnotesize Article 5 (1) A minor must obtain the consent of the minor's ...}};
% Text Node
\draw (69,478.91) node [anchor=north west][inner sep=0.75pt]   [align=left] {{\footnotesize explanation: $\displaystyle \langle keyword:"minor",relevant:"juridical\ act"\rangle $}};

\end{tikzpicture}
    }
    \caption{LoT Prompting in Japanese Civil Law Retrieval Setup}
    \label{fig:diagram-exp1}
\end{figure}
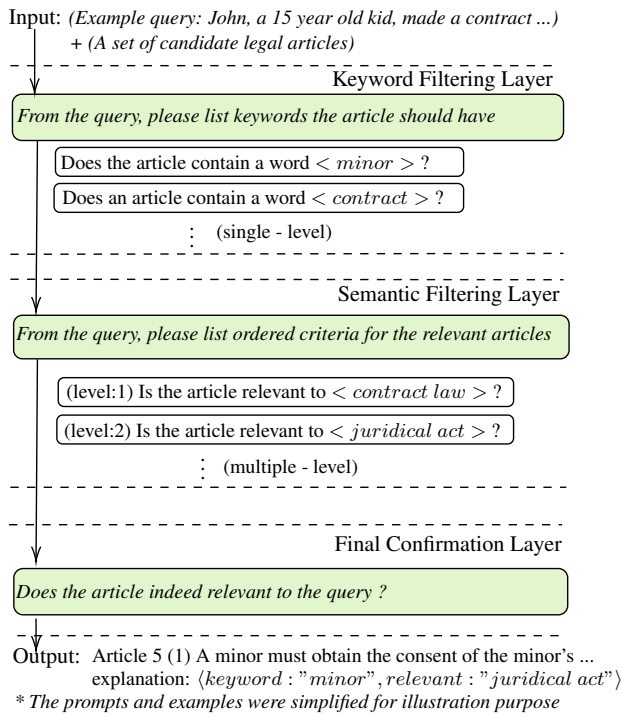

The proposed design contains multiple layers following the idea of LoT (as demonstrated in Figure \ref{fig:diagram-exp1}):

\begin{enumerate}
    \item \textbf{Keyword Filtering Layer (KFL):} The first layer involves keyword filtering. The keywords are suggested by a Large Language Model (LLM) after it analyzes the query. This layer uses \texttt{at-least-1} metric.
   
    \item \textbf{Semantic Filtering Layer (SFL):} Following the keyword filtering, the semantic filtering layer applies conditions suggested by the LLM to the selected candidates. This is a multiple-level layer, organized from general to specific conditions, helping refine the selection progressively. This layer uses \texttt{max-count} metric.

    \item \textbf{Final Confirmation Layer (FCL):} In the final layer, the original query is used once again. The LLM is asked to confirm whether the candidate articles can indeed help answer the query. This step ensures that only the most relevant articles are selected. This layer uses \texttt{all} metric.
\end{enumerate}

Besides, we also challenge the proposed design with two simpler settings. In the first setting, the \textbf{SFL} is omitted, meaning that candidates are directly evaluated against the original query immediately after keyword filtering. In the second, even more minimal setting, we remove the \textbf{SFL} and \textbf{KFL} entirely.

\begin{table*}[ht]
\centering
\begin{tabular}{|l|c|c|c|}
\hline
\textbf{System} & \textbf{Precision} & \textbf{Recall} & \textbf{F2} \\
\hline
Proposed Design following LoT &0.838 &	0.839 &	0.835 \\
\hline
Direct Validation with Query (w. KFL) & 0.546	& 0.853 &	0.563 \\
\hline
Direct Validation with Query (w/o. KFL) & 0.432  &	0.885&	0.449 \\
\hline
\multicolumn{4}{|c|}{\textit{Other Comparative Models}} \\
\hline
JNLP \cite{nguyen2024pushing} & 0.709 & 0.870 & 0.807 \\
\hline
CAPTAIN \cite{nguyen2024captain} & 0.732 & 0.845 & 0.800 \\
\hline
TQM \cite{li2024towards} & 0.785 & 0.800 & 0.782 \\
\hline
NOWJ \cite{nguyen2024nowj} & 0.690 & 0.835 & 0.772 \\
\hline
AMHR \cite{nighojkar2024amhr}& 0.651 & 0.825 & 0.749 \\
\hline
UA \cite{housam2024legal} & 0.610 & 0.800 & 0.711 \\
\hline
% BM24 & 0.282 & 0.795 & 0.539 \\
% \hline
% PSI & 0.090 & 0.085 & 0.086 \\
% \hline
\end{tabular}
\caption{Comparison of Different Systems in Terms of Precision, Recall, and F2}
\label{table:comparison}
\end{table*}

Upon examining the results in Table~\ref{table:comparison}, several insights can be drawn.
The \textbf{Proposed Design following LoT} demonstrated the highest F2 score (0.835), outperforming the best systems in COLIEE 2024. This high F2 score indicates a well-balanced performance in terms of precision (0.838) and recall (0.839), showcasing the effectiveness of the Layer-of-Thoughts (LoT) approach. The structured filtering layers, including keyword and semantic filtering, ensure a balanced retrieval of relevant articles, minimizing false positives while maximizing true positives.

In contrast, the two direct validation systems exhibited higher recall but significantly lower precision. The \textbf{Direct Validation with Query (w. KFL)} system achieved a recall of 0.853 but only a precision of 0.546, resulting in an F2 score of 0.563. The \textbf{Direct Validation without Query (w/o. KFL)} system had a recall of 0.885 but a precision of merely 0.432, leading to an F2 score of 0.449. These results suggest a more tolerant approach, where the absence of semantic filtering layer causes an influx of irrelevant articles, thus lowering precision. An extreme instance highlighted this issue, where the LLM marked up to 400 articles as relevant while the gold standard identified only 2. This indicates a high risk of false positives in these settings. Combining filtering by keywords can be a simple yet effective strategy to reduce false positives. 

Among the comparative models, \textbf{JNLP} achieved the highest F2 score (0.807) with good precision (0.709) and recall (0.870), followed by \textbf{CAPTAIN} with an F2 score of 0.800 (precision 0.732, recall 0.845). \textbf{TQM} also performed well with an F2 score of 0.782 (precision 0.785, recall 0.800). While these models showed strong overall performance, they did not surpass the balance achieved by the proposed design.

Other models such as \textbf{NOWJ}, \textbf{AMHR}, and \textbf{UA} displayed competent results with F2 scores of 0.772, 0.749, and 0.711, respectively. These scores indicate an effective yet slightly imbalanced performance compared to the top-performing models and the proposed design.

The high F2 score of the proposed design illustrates its superior balance between precision and recall, unlike the direct validation systems, which prioritize recall at the expense of precision. This balance is crucial in practical scenarios to ensure relevant legal articles are retrieved accurately, minimizing the effort required for further verification.

\subsection{Normative Sentence Retrieval}
In the context of automated driving, it is essential to have a comprehensive understanding of both explicit and implicit traffic rules. While explicit rules are often codified in statutes and regulations, implicit rules are frequently derived from judicial decisions and case law. These implicit rules, which are crucial for ensuring safe and lawful driving behavior, need to be systematically identified and made explicit. 

Extracting normative sentences from court decisions is vital for developing a comprehensive corpus of traffic rules for automated driving systems. Advanced natural language processing (NLP) techniques, machine learning models, and expert validation can be employed to identify and articulate implicit rules, thereby enhancing the traffic regulation framework. This ensures that automated driving systems operate under a robust and legally sound set of traffic norms, promoting safety and compliance.
%This section outlines the methodology for retrieving normative sentences from court decisions to enhance the body of traffic rules for automated driving systems.
%Autonomous driving is a transformative technology set to revolutionize transportation systems worldwide. A major challenge is ensuring that autonomous vehicles can effectively comply with the complex traffic rules governing road safety and efficiency. Addressing this challenge requires a deep understanding of how machines can interpret, process, and adhere to traffic regulations. Ensuring compliance involves a comprehensive approach that integrates logic programming, machine learning, rule mining from case law, and consideration of implicit rules and general clauses. This approach helps autonomous vehicles navigate diverse traffic scenarios while prioritizing safety, legal compliance, and efficient traffic flow.
%\subsubsection{Datasets:}

Currently, there are no annotated datasets available to train a model for retrieving normative sentences specifically related to Section 6 of the German Road Traffic Regulations (i.e., § 6 StVO)\footnote{\url{https://dejure.org/gesetze/StVO/6.html}}. Normative sentences are those identified as additions, clarifications, or interpretations of the traffic rule stipulated by § 6 StVO. Developing rules or patterns to accurately identify and differentiate these sentences is challenging because not every normative sentence is our focus; the sentence must be both normative and related to § 6 StVO. This challenge arises to distinguish between the broader category of normative sentences and those specifically pertinent to § 6 StVO. The complexity lies in the dual criteria of normativeness and relevance, which necessitates a more sophisticated approach than mere keyword matching or rule-based retrieval systems.

\begin{figure}[ht]
    \centering
    \includegraphics[width=.5\textwidth]{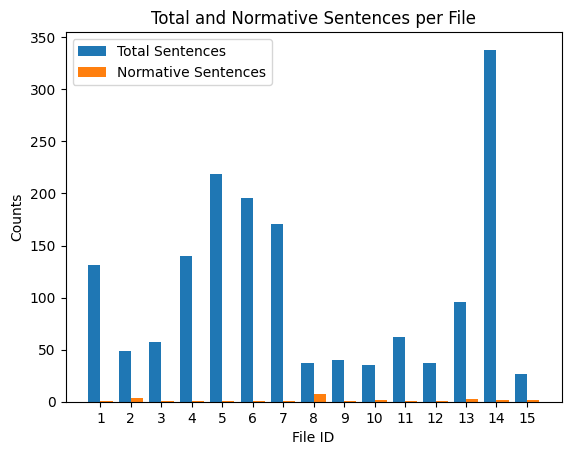}
    \caption{Counts of Total and Annotated Normative Sentences per File}
    \label{fig:analysis}
\end{figure}

The evaluation dataset used in this study consists of English translations of 15 court decisions pertinent to § 6 StVO. We randomly selected these decisions from the \textit{Case law on Section 6 StVO}\footnote{\url{https://dejure.org/dienste/lex/StVO/6/1.html}} and translated them into English. The focus is on the reasoning sections of these decisions, as they are most pertinent for identifying normative statements. The dataset excludes other sections that do not contribute to understanding the normative content.
Figure \ref{fig:analysis} shows a graphical representation of the total sentences and the normative sentences for each file ID.

% The number of total sentences and the number of human-annotated gold standard normative sentences for each file are presented in Table~\ref{table:NormativeSentenceDataset}. 
% \begin{table*}[ht]
% \centering
% \begin{tabular}{|c|c|c|}
% \hline
% \textbf{File ID} & \textbf{Total Sentences} & \textbf{Gold Normative Sentences} \\
% \hline
% 1 &131 &1 \\
% \hline
% 2 &49 &4 \\
% \hline
% 3 &57 &1 \\
% \hline
% 4 &140 &1 \\
% \hline
% 5 &219 &1 \\
% \hline
% 6 &196 &1 \\
% \hline
% 7 &171 &1 \\
% \hline
% 8 &37 &7 \\
% \hline
% 9 &40 &1 \\
% \hline
% 10 &35 &2 \\
% \hline
% 11 &62 &1 \\
% \hline
% 12 &37 &1 \\
% \hline
% 13 &96 &3 \\
% \hline
% 14 &338 &2 \\
% \hline
% 15 &27 &2 \\
% \hline
% \end{tabular}
% \caption{Counts of Total and Annotated Normative Sentences}
% \label{table:NormativeSentenceDataset}
% \end{table*}
%\subsubsection{Evaluation Metrics:}

Minimizing manual effort in identifying normative sentences is a key objective. For each input file, if the system extracts 10 sentences and includes all normative sentences among them, we consider this an optimal outcome because it balances efficiency and comprehensiveness. Extracting a manageable number of sentences while ensuring all normative ones are included allows for quick verification and reduces the burden on manual review. Therefore, our evaluation metrics prioritize achieving higher recall to ensure that most, if not all, normative sentences are captured.
%\subsubsection{Implementation Details:}

\begin{figure}
    \centering
    \scalebox{0.7}{
        \tikzset{every picture/.style={line width=0.75pt}} %set default line width to 0.75pt        

\begin{tikzpicture}[x=0.75pt,y=0.75pt,yscale=-1,xscale=1]
%uncomment if require: \path (0,566); %set diagram left start at 0, and has height of 566

%Straight Lines [id:da35385755314536205] 
\draw    (31.52,23.65) -- (31.52,65.65) ;
\draw [shift={(31.52,67.65)}, rotate = 270] [color={rgb, 255:red, 0; green, 0; blue, 0 }  ][line width=0.75]    (10.93,-3.29) .. controls (6.95,-1.4) and (3.31,-0.3) .. (0,0) .. controls (3.31,0.3) and (6.95,1.4) .. (10.93,3.29)   ;
%Rounded Rect [id:dp3781674363270149] 
\draw   (90.47,112.12) .. controls (90.47,109.82) and (92.33,107.96) .. (94.63,107.96) -- (352.84,107.96) .. controls (355.14,107.96) and (357,109.82) .. (357,112.12) -- (357,124.6) .. controls (357,126.9) and (355.14,128.76) .. (352.84,128.76) -- (94.63,128.76) .. controls (92.33,128.76) and (90.47,126.9) .. (90.47,124.6) -- cycle ;
%Rounded Rect [id:dp5703080499776279] 
\draw  [fill={rgb, 255:red, 184; green, 233; blue, 134 }  ,fill opacity=0.4 ] (13.47,75.94) .. controls (13.47,72.65) and (16.13,69.98) .. (19.42,69.98) -- (358.51,69.98) .. controls (361.8,69.98) and (364.47,72.65) .. (364.47,75.94) -- (364.47,93.8) .. controls (364.47,97.09) and (361.8,99.76) .. (358.51,99.76) -- (19.42,99.76) .. controls (16.13,99.76) and (13.47,97.09) .. (13.47,93.8) -- cycle ;
%Straight Lines [id:da875127085120859] 
\draw  [dash pattern={on 4.5pt off 4.5pt}]  (9,45) -- (388.47,44.76) ;
%Straight Lines [id:da6664986440018623] 
\draw  [dash pattern={on 4.5pt off 4.5pt}]  (13,184) -- (392.47,182.76) ;
%Straight Lines [id:da4512203005524107] 
\draw    (32.47,101.76) -- (31.48,223.76) ;
\draw [shift={(31.47,225.76)}, rotate = 270.46] [color={rgb, 255:red, 0; green, 0; blue, 0 }  ][line width=0.75]    (10.93,-3.29) .. controls (6.95,-1.4) and (3.31,-0.3) .. (0,0) .. controls (3.31,0.3) and (6.95,1.4) .. (10.93,3.29)   ;
%Straight Lines [id:da23299907274705278] 
\draw  [dash pattern={on 4.5pt off 4.5pt}]  (13,202) -- (390.47,200.76) ;
%Straight Lines [id:da939804401572238] 
\draw  [dash pattern={on 4.5pt off 4.5pt}]  (14,273) -- (392.47,271.76) ;
%Straight Lines [id:da11504731618525965] 
\draw    (30.47,262.76) -- (30.47,302.65) ;
\draw [shift={(30.47,304.65)}, rotate = 270] [color={rgb, 255:red, 0; green, 0; blue, 0 }  ][line width=0.75]    (10.93,-3.29) .. controls (6.95,-1.4) and (3.31,-0.3) .. (0,0) .. controls (3.31,0.3) and (6.95,1.4) .. (10.93,3.29)   ;
%Rounded Rect [id:dp04350494164087504] 
\draw  [fill={rgb, 255:red, 184; green, 233; blue, 134 }  ,fill opacity=0.4 ] (15.47,312.27) .. controls (15.47,308.59) and (18.45,305.6) .. (22.14,305.6) -- (356.85,305.6) .. controls (360.53,305.6) and (363.52,308.59) .. (363.52,312.27) -- (363.52,332.29) .. controls (363.52,335.97) and (360.53,338.96) .. (356.85,338.96) -- (22.14,338.96) .. controls (18.45,338.96) and (15.47,335.97) .. (15.47,332.29) -- cycle ;
%Straight Lines [id:da32384875636368626] 
\draw  [dash pattern={on 4.5pt off 4.5pt}]  (12,282) -- (388.47,281.76) ;
%Straight Lines [id:da8612806227614322] 
\draw  [dash pattern={on 4.5pt off 4.5pt}]  (17,351) -- (385.47,349.76) ;
%Straight Lines [id:da4133987311781424] 
\draw    (30.47,339.69) -- (30.51,377.65) ;
\draw [shift={(30.52,379.65)}, rotate = 269.93] [color={rgb, 255:red, 0; green, 0; blue, 0 }  ][line width=0.75]    (10.93,-3.29) .. controls (6.95,-1.4) and (3.31,-0.3) .. (0,0) .. controls (3.31,0.3) and (6.95,1.4) .. (10.93,3.29)   ;
%Rounded Rect [id:dp9531705685262362] 
\draw   (89.47,140.12) .. controls (89.47,137.82) and (91.33,135.96) .. (93.63,135.96) -- (352.84,135.96) .. controls (355.14,135.96) and (357,137.82) .. (357,140.12) -- (357,152.6) .. controls (357,154.9) and (355.14,156.76) .. (352.84,156.76) -- (93.63,156.76) .. controls (91.33,156.76) and (89.47,154.9) .. (89.47,152.6) -- cycle ;
%Rounded Rect [id:dp19902547409146187] 
\draw  [fill={rgb, 255:red, 184; green, 233; blue, 134 }  ,fill opacity=0.4 ] (13.33,234.34) .. controls (13.33,230.27) and (16.62,226.98) .. (20.68,226.98) -- (354.16,226.98) .. controls (358.22,226.98) and (361.52,230.27) .. (361.52,234.34) -- (361.52,256.4) .. controls (361.52,260.47) and (358.22,263.76) .. (354.16,263.76) -- (20.68,263.76) .. controls (16.62,263.76) and (13.33,260.47) .. (13.33,256.4) -- cycle ;

% Text Node
\draw (8,3) node [anchor=north west][inner sep=0.75pt]   [align=left] {Input: {\footnotesize \textit{(Example traffic rule (§ 6 StVO): Anyone wishing to pass \ a .....)}}};
% Text Node
\draw (17,380) node [anchor=north west][inner sep=0.75pt]   [align=left] {Output: {\footnotesize According to this provision, a road user who wishes to pass }\\{\footnotesize  \ \ \ \ \ \ \ \ \ \ \ \ \ \ \ a stopped vehicle, a barrier or any other obstacle .........}};
% Text Node
\draw (197,158.66) node [anchor=north west][inner sep=0.75pt]   [align=left] {$\displaystyle \vdots $};
% Text Node
\draw (175,205) node [anchor=north west][inner sep=0.75pt]  [font=\normalsize] [align=left] {Normativeness classification layer};
% Text Node
\draw (214.65,162.63) node [anchor=north west][inner sep=0.75pt]  [font=\footnotesize] [align=left] {(single - level)};
% Text Node
\draw (225,288) node [anchor=north west][inner sep=0.75pt]  [font=\normalsize] [align=left] {Final Confirmation Layer};
% Text Node
\draw (18.14,76.6) node [anchor=north west][inner sep=0.75pt]   [align=left] {{\footnotesize \textit{From traffic rule, please list keywords the sentence should have }}};
% Text Node
\draw (229,46) node [anchor=north west][inner sep=0.75pt]  [font=\normalsize] [align=left] {Keyword Filtering Layer};
% Text Node
\draw (96.63,110.96) node [anchor=north west][inner sep=0.75pt]  [font=\footnotesize] [align=left] {Does the sentence contain a word $\displaystyle < pass >$ ?};
% Text Node

\draw (97.03,139.96) node [anchor=north west][inner sep=0.75pt]  [font=\footnotesize] [align=left] {Does the sentence contain a word $\displaystyle < allow >$ ?};
% Text Node
\draw (22.68,229.98) node [anchor=north west][inner sep=0.75pt]  [font=\footnotesize] [align=left] {\textit{Please assess the degree of traffic rule related normativeness for} \\\textit{each sentence on a scale from 0 to 100.}};
% Text Node
\draw (24.14,308.6) node [anchor=north west][inner sep=0.75pt]   [align=left] {{\footnotesize \textit{Does the sentence add to, clarify, or interpret § 6 StVO?}}\\{\footnotesize \textit{Rank sentences by relevance to § 6 StVO and extract the top 10.}}};
% Text Node
\draw (46,22) node [anchor=north west][inner sep=0.75pt]   [align=left] {{\footnotesize \textit{+ (A complete list of sentences from the court decision)}}};
% Text Node
\draw (14,461) node [anchor=north west][inner sep=0.75pt]   [align=left] {{\footnotesize \textit{* The prompts and examples were simplified for illustration purpose }}};
% Text Node
\draw (68,409.91) node [anchor=north west][inner sep=0.75pt]   [align=left] {{\footnotesize explanation: $\displaystyle  \begin{array}{l}
 \langle  keyword: ``pass, obstacle, ...", \\
degreeOfNormativeness: ``90",\ \\
relevant: ``yes\ ( interpret) ;top-1"\rangle 
\end{array}$}};

\end{tikzpicture}
    }
    \caption{LoT Prompting in Normative Sentence Retrieval Setup}
    \label{fig:diagram-exp2}
\end{figure}

\begin{table*}[ht]
\centering
\begin{tabular}{|l|c|c|c|}
\hline
\textbf{System} & \textbf{Precision} & \textbf{Recall} & \textbf{F2} \\
\hline
Proposed Design following LoT &0.187 &	0.966 &	0.527 \\
\hline
\multicolumn{4}{|c|}{\textit{Other Comparative Models}} \\
\hline
Chain-of-Thoughts & 0.167 & 0.862 & 0.470 \\
\hline
BM25 Retriever& 0.153 & 0.793 & 0.432 \\
\hline
Retrieve \& Re-Rank & 0.140 & 0.724 & 0.395 \\
\hline
\end{tabular}
\caption{Comparison of Normative Sentence Retrieval Systems in Terms of Precision, Recall, and F2}
\label{table:comparisonOfNormativeSentenceRetrieval}
\end{table*}

A multi-layered approach using LoT Prompting is proposed for extracting normative sentences, as demonstrated in Figure \ref{fig:diagram-exp2}. 
\begin{enumerate}
    \item \textbf{Keyword Filtering Layer (KFL):} The first layer involves using a Large Language Model (LLM) to generate a list of relevant keywords associated with traffic rules. To ensure comprehensive coverage, word embedding technique, such as Word2Vec, is applied to identify and match keywords or their closely related terms within sentences. This layer uses \texttt{at-least-2} metric, that is sentences containing at least two keywords that are either exact matches or semantically similar are selected for further processing. 
    \item \textbf{Normativeness Classification Layer (NCL):} Next, the second layer prompts the LLM to evaluate the filtered sentences. Each sentence is assessed based on its degree of normativeness, with normative sentences defined as those having a normativeness score of 70 or higher.
    \item \textbf{Final Confirmation Layer (FCL):} The final layer involves the LLM verifying the relevance of the normative sentences to § 6 StVO. Each candidate sentence is cross-referenced with the specific description of § 6 StVO to ensure its relevance. This layer uses \texttt{rank-weight} metric to refine the selection process by narrowing down the results to the top 10 most relevant sentences, thereby focusing on the most pertinent information.
\end{enumerate}
In this study, we employed GPT-4o as the LLM with parameters configured as follows: \texttt{temperature} at \texttt{0}, \texttt{top-p} at \texttt{1}, \texttt{frequency penalty} at \texttt{0}, and \texttt{presence penalty} at \texttt{0}.

Table~\ref{table:comparisonOfNormativeSentenceRetrieval}
% Figure~\ref{fig:analysis}
presents the results. Comparisons are made with three baseline methods: BM25~\cite{robertson1994some}, Retrieve \& Re-Rank\footnote{\url{https://sbert.net/examples/applications/retrieve_rerank/README.html}}, and the Chain-Of-Thoughts approach. BM25 is a widely used ranking function in information retrieval that measures the relevance of documents (or sentence, in this experiment) to a query based on term frequency and document length. Retrieve \& Re-Rank involves an initial retrieval phase to identify a set of candidate sentences and a subsequent re-ranking phase to improve their relevance. In the experiment, the English translation of § 6 StVO serves as the query. The Chain-Of-Thoughts approach involves reasoning through sequences of thought or processing steps to extract information. Comparing with these methods helps evaluate the proposed approach's performance against established retrieval techniques and advanced reasoning models. Given that the number of normative sentences per file ranges from 1 to 7 (as indicated by 
% Table~\ref{table:NormativeSentenceDataset}
Figure~\ref{fig:analysis}
), focusing on extracting the top 10 sentences is justified. The top-k value is set to 10 for all models. The LoT method demonstrates precise and reliable extraction of normative sentences, achieving a higher recall score compared to the baselines. Although precision and F2 scores are not exceptionally high, mainly because most files contain only one gold standard normative sentence while our method extracts ten sentences, the LoT method still outperforms the baseline approaches.

\section{Discussion}
The experimental results highlight the merits and potential drawbacks of the Layer-of-Thoughts (LoT) Prompting in various information retrieval tasks. For Japanese Civil Law retrieval, the LoT approach consistently outperformed traditional methods and even advanced retrieval models used in the COLIEE competition. This demonstrates that the integration of hierarchical thoughts and structured reasoning can significantly enhance the quality of retrieved documents by balancing both precision and recall. The high F2 scores indicate that LoT Prompting effectively captures the nuances of the task, leading to a more precise and well-rounded retrieval process.

In the context of normative sentence retrieval, the LoT method also showed promising results. Despite the inherently challenging nature of this task, which involves high variability and implicit rules, LoT managed to achieve a recall score close to 1.0. This high recall is particularly valuable in legal contexts, where missing relevant sentences could lead to critical oversights. However, the precision scores indicate that while most relevant sentences are captured, further refinement is needed to reduce false positives and enhance the overall relevance of the extracted sentences.

The scalability and efficiency of the LoT Prompting are largely supported by its hierarchical structure. The use of strength metrics to prioritize thoughts ensures that the computational overhead is managed effectively. By leveraging constraint hierarchies, LoT Prompting efficiently narrows down the search space, making it well-suited for large-scale information retrieval tasks. This structured approach not only reduces unnecessary computations but also improves the speed and accuracy of retrieving relevant documents.

One of the standout features of the LoT Prompting is its explainability. By breaking down the retrieval process into hierarchical levels of thoughts, LoT Prompting provides clear and interpretable pathways from the query to the final retrieved documents. Each step in the hierarchy can be traced and understood, offering transparency in how decisions are made. This interpretability is invaluable, particularly in legal and regulatory contexts, where the rationale behind document retrieval must be clear and defensible.

\section{Conclusion}

In this paper, we introduced the Layer-of-Thoughts (LoT) Prompting, which incorporates constraint hierarchies to enhance the retrieval of information through structured prompting. Our experiments on Japanese Civil Law retrieval and normative sentence extraction tasks demonstrated that the LoT Prompting significantly improves both precision and recall compared to baseline and state-of-the-art methods. The ability of LoT to balance these metrics while providing explainable and efficient retrieval processes makes it a compelling tool for complex information retrieval tasks. Future work will focus on further refining LoT Prompting to enhance precision, particularly in tasks involving highly variable and implicit information.

\subsubsection*{Acknowledgements.} The authors would like to thank Prof. Dr. Georg Borges and Carmen Martin for providing the evaluation dataset used in the Normative Sentence Retrieval. This research was supported by \textquotedblleft Strategic Research Projects\textquotedblright grant from ROIS
(Research Organization of Information and Systems), JSPS KAKENHI Grant Numbers,
JP22H00543, JST, AIP Trilateral AI Research, Grant Number JPMJCR20G4,
and the MEXT \textquotedblleft R\&D Hub Aimed at Ensuring Transparency and Reliability
of Generative AI Models\textquotedblright project.
% \bibliographystyle{plain}
% \bibliography{references}

%% The file kr.bst is a bibliography style file for BibTeX 0.99c
\bibliographystyle{kr}
\bibliography{references}

\end{document}